\documentclass[twocolumn,aps,pre,superscriptaddress,reprint]{revtex4-1}
\usepackage{amsmath}
\usepackage{amssymb}
\usepackage{graphicx}

\usepackage[utf8]{inputenc}

\usepackage{subfigure}          
\usepackage{url}                
\usepackage{color}              

\newcommand{\e}[1]{\times 10^{#1}}
\newcommand{\pf}[1]{\texttt{#1}} 

\begin{document}

\title{Using a reservoir computer to learn chaotic attractors, with applications to chaos synchronisation and cryptography}
\author{Piotr Antonik}
\thanks{piotr.antonik@centralesupelec.fr}
\thanks{Note: P.A. and M.G. contributed equally to this work.}
\affiliation{CentraleSup\'elec, Campus de Metz, Universit\'e Paris Saclay, F-57070 Metz, France}
\affiliation{LMOPS EA 4423 Laboratory, CentraleSup\'elec \& Universit\'e de Lorraine, F-57070 Metz, France}
\author{Marvyn Gulina}
\affiliation{Namur Institute for Complex Systems, Universit\'e de Namur, B-5000 Namur, Belgium}
\author{Ja\"{e}l Pauwels}
\affiliation{Applied Physics Research Group, Vrije Universiteit Brussels, B-1050 Brussels, Belgium}
\affiliation{Laboratoire d'Information Quantique, Universit\'e libre de Bruxelles, B-1050 Brussels, Belgium}
\author{Serge Massar${}^5$}

\begin{abstract}
  Using the machine learning approach known as reservoir computing, it is possible to train one dynamical system to emulate another. 
  We show that such trained reservoir computers reproduce the properties of the attractor of the chaotic system sufficiently well to exhibit chaos synchronisation.
  That is, the trained reservoir computer, weakly driven by the chaotic system, will synchronise with the chaotic system. Conversely, the chaotic system, weakly driven by a trained reservoir computer, will synchronise with the reservoir computer. We illustrate this behaviour on the Mackey-Glass and Lorenz systems. We then show that trained reservoir computers can be used to crack chaos based cryptography and illustrate this on a chaos cryptosystem based on the Mackey-Glass system. 
  We conclude by discussing why reservoir computers are so good at emulating chaotic systems.
\end{abstract}

\maketitle

\section{Introduction}

Can one train a nonlinear dynamical system to emulate a different nonlinear chaotic dynamical system? This question has been answered positively in \cite{jaeger2004harnessing,antonik2017brain,lu2017reservoir,pathak2017using,pathak2018model} in the context of the machine learning technique known as reservoir computing. But much remains to be learned concerning the quality of this emulation.

Reservoir Computing (RC) \cite{jaeger2001echo,maass2002real,jaeger2004harnessing,verstraeten2007experimental}, on which this approach is based, is a machine learning technique in which a nonlinear dynamical system with a large number of internal nodes (called the reservoir) is driven by a time dependent signal. The connections between the internal variables are chosen at random and then kept fixed (except, possibly, for a few global parameters that may be adjusted). In many implementations, the reservoir is  a recurrent neural network  with fixed connections. The output of the reservoir computer is a single node whose state is given by a linear combination of the states of the internal variables. The weights of this linear combination are trained to match the output as closely as possible to a desired target. Although conceptually simple, reservoir computing is powerful enough to equal other algorithms on hard tasks such as channel equalisation, phoneme recognition, and others (see \cite{lukovsevivcius2009survey,lukovsevivcius2012reservoir} for reviews).

The theory of reservoir computing is not very advanced (the situation being similar to many machine learning approaches which work well in practice but lack formal explanations for their performance). One of the most useful theoretical concepts is that of the linear and nonlinear memory capacity of reservoirs, see \cite{jaeger2001short,dambre2012information,marzen2017difference,inubushi2017reservoir}. It was also shown recently that a variant of reservoir computer is universal in the category of fading memory filters \cite{grigoryeva2017universal}.

Reservoir computing has also been implemented experimentally with performance comparable to digital implementations \cite{appeltant2011information,larger2012photonic,paquot2012optoelectronic} with photonic implementations presenting particularly high speeds \cite{brunner2012parallel,vandoorne2014experimental,Larger2017high}, see \cite{van2017advances} for a review.

In the case where one wants the reservoir computer to emulate a dynamical system, the reservoir is first driven by the state of the dynamical system, and trained to predict this state one timestep in the future. After training one closes the loop and feeds the output of the reservoir back into itself, whereupon it will develop autonomous dynamics that are -- one hopes -- close to those of the original dynamical system.

This approach was originally introduced to forecast the trajectories of chaotic dynamical systems, where it reached record forecasting horizons \cite{jaeger2004harnessing}. These results were improved recently in \cite{grigoryeva2017universal}. 
In addition this method was used numerically in \cite{lu2017reservoir} to infer the values of hidden degrees of freedom of the dynamical system, in \cite{pathak2017using} to estimate its Lyapunov exponents, and in \cite{pathak2018model} to predict spatio-temporal chaos. It  was also implemented in an opto-electronic system \cite{antonik2017brain} where it was shown  that the experimental reservoir could be trained to have similar dynamics to the original system (similar spectrum, Lyapunov exponents, etc).
From these works it is clear that a  reservoir computer trained as described above can emulate another, a priori completely different (possibly chaotic) dynamical system. However much remains to be understood about the quality and accuracy of the method, as well as its potential limitations. Here we show how trained reservoir computers can be used to replace dynamical systems in two other applications: chaos synchronisation and cracking chaos-based cryptography.

One of the most surprising aspects of chaos theory is the synchronisation of two identical chaotic systems. Although the dynamics of each system separately is unpredictable, if one system is driven by the other, the two systems will synchronise \cite{pecora1990synchronization,pecora1991driving}. This phenomenon has been extensively studied, see e.g. the review \cite{boccaletti2002synchronization}.

In the first part of this work, we show that a reservoir computer, trained to emulate a chaotic system as described above, if driven by the original system, will synchronise with it. This demonstrates that not only is the dynamics of the reservoir computer emulator superficially similar to that of the original system, but that its chaotic attractor has similar properties. That is, it captures a large part of the characteristics of the dynamics of the original system. We illustrate this on two examples, the Lorenz \cite{lorenz1963deterministic} and Mackey-Glass systems \cite{mackey1977oscillation}.

After the discovery of chaos synchronisation, considerable effort was devoted to trying to use this effect and the unpredictability of chaotic systems to hide secret messages, see e.g. \cite{parlitz1992transmission,cuomo1993circuit,colet1994digital}. A series of experimental demonstrations were realised \cite{vanwiggeren1998communication,goedgebuer1998optical,argyris2005chaos}. However, it was later shown that chaos-based cryptography is fundamentally insecure, as there are efficient ways for an eavesdropper to find the parameters of the chaotic system (which play the role of secret key), at least using plain text attack. This was first demonstrated on a series of examples, see e.g. \cite{ponomarenko2002extracting,udaltsov2003cracking,wang2004error,alvarez2004cryptanalyzing,li2005breaking,alvarez2005breaking,prokhorov2008encryption}, and then in full generality \cite{anstett2006chaotic}.

As an application of our results on chaos synchronisation we consider using the reservoir computer to crack chaos-based cryptography. We will study a cryptosystem based on the Mack-Glass equations previously studied from the point of view of encryption and cryptanalysis in \cite{ponomarenko2002extracting,prokhorov2008encryption}. This is an ''open loop'' configuration of the type widely used in experiments because of its robustness. Our attack based on reservoir computing performs similarly to the parameter search studied in \cite{ponomarenko2002extracting}.

We conclude this article with a general discussion of why reservoir computers are able to crack chaos-based cryptosystems. This should be viewed as a general discussion which shows the plausibility of this kind of attack, but without any claim to mathematical rigour or an understanding of the efficiency of such attacks.

\section{Reservoir computing}
\label{sec:rc}

\subsection{Basic principles}

The reservoir computer used in this work is a discrete-time echo state network, as introduced in \cite{jaeger2004harnessing,jaeger2001echo}. The reservoir states vector $x$, consisting of $N$ neurons, is updated following the equation
\begin{align}
  x(n) &= (1- C a) x(n-1) \nonumber \\
  & + C \tanh \left( w_\text{in} u(n) + W x(n-1) + w_\text{back} d(n-1) \right),
  \label{eq:rc}
\end{align}
where $n\in\mathbb{Z}$ is the discrete time, $C$ is a timescale constant, $a$ is the leak rate, $W$ is a $N\times N$ matrix of internal connection weights, $w_\text{back}$ is $N$-size weight vector for feedback connections from the output to the reservoir and $w_\text{in}$ is a $N$-size vector and $u$ is a constant. Together, $C$ and $a$ realise a low-pass filter with adjustable properties.

The elements of $w_\text{in}$, $W$ and $w_\text{back}$ are chosen from a uniform distribution over the interval $[-1,+1]$. A reservoir computer must be not too far from the edge of chaos to exhibit good performance. To this end the matrix $W$ is then rescaled to adapt its spectral radius. The vectors  $w_\text{in}$ and $w_\text{back}$ are possibly also rescaled to adapt the strength of the input and feedback. Throughout this work the input bias is fixed to $u=0.2$.
To fix the other parameters of the reservoir computer, such as $C$, $a$, spectral radius of $W$, scaling of $w_\text{in}$ and $w_\text{back}$ we took inspiration from  \cite{jaeger2001echo}.
When the parameter values were not available for a specific task, suitable values were chosen heuristically, but without carrying out a systematic search over all possible values. The chosen values of parameters are given in the text below.

The output equation of a single-output network is given by a dot product
\begin{equation}
  y(n) = w_\text{out} \left[ x(n), u(n) \right],
  \label{eq:rcout}
\end{equation}
where $\left[ x(n), u(n) \right]$ is the concatenation of the reservoir states vector $x(n)$ with the input $u(n)$, and $w_\text{out}$ are $N+1$ output weights (also known as the output mask). 

During training we adjust the weights $w_\text{out}$ so that the output $y(n)$ is as close as possible to the desired output $\tilde y(n)$.
To this end we minimise the Normalised Mean Square Error (NMSE), given by
\begin{equation}
  \text{NMSE} = \frac{\left\langle \left( y(n) - \tilde y(n) \right)^2 \right\rangle}{\left\langle \left( \tilde y(n) - \langle \tilde y(n) \rangle \right)^2 \right\rangle}\ .
  \label{eq:mse}
\end{equation}
The NMSE indicates how far the time series $y(n)$ generated by the reservoir deviates from the target  time series $\tilde y(n)$. The resulting value is  straightforward to interpret: $\text{NMSE}=0$ means that the two series match, while $\text{NMSE}=1$ indicates no similarity at all. 
Minimising the NMSE with respect to the readout weights gives rise to a system of linear equations that is readily solved. 
We do not use ridge regression \cite{tikhonov1995numerical} (except in the Lorenz task, see below) as there is enough training data to avoid overfitting.

After training we evaluate the performance of the reservoir on a new data set. When we report NMSE values, it is the values evaluated on the test sequence.

In the present work, the reservoir computer must predict or process continuous time signals $u(t)$. To pass from continuous to discrete time, we sample the continuous time input as
\begin{equation}
u(n)=u(t+n\Delta)
\label{eq:sample}
\end{equation}
where $\Delta$ is the sampling interval. For the different applications, we quote the sampling interval used.

\subsection{Training to crack chaotic cryptography}
\label{SubSec:TrainCrypto}

When training to crack chaotic cryptography, we operate the reservoir as described above. We set $w_\text{back}=0$. We consider a plain text attack in which Eve has access to the signal sent by Alice to Bob, and to the message that we encrypted by Alice. For training we take $u(n)$ to be the signal intercepted by Eve. We take $\tilde y(n)=m(n)$ to be the message encoded by Alice. During the training phase the weights 
$w_\text{out}$ are thus adjusted so that the reservoir outputs the encrypted message.
Eve's reservoir computer is then ready to decrypt new encrypted messages.

\subsection{Training to emulate chaotic systems}

When training the reservoir to emulate a chaotic system, for instance, for chaos synchronisation, we proceed as follows. Denote by $s(n)$ the time series of the chaotic system we wish to emulate.

We set the input to a constant
\begin{equation}
\label{eq:ucst}
  u(n)=0.2.
\end{equation}

During training  Eqs. \ref{eq:rc}, \ref{eq:rcout}, \ref{eq:mse}, \ref{eq:ucst} are supplemented by
\begin{eqnarray}
  d(n)&=&s(n)    \hspace{1cm} \text{(during training)}, \nonumber\\
  \tilde y(n)&=&s(n)
  \hspace{1cm} \text{(during training)}.
  \label{eq:rcTrain}
\end{eqnarray}
That is, the training phase is  used to optimise the readout weights $w_\text{out}$ so that the reservoir predicts the next point $s(n)$ in the input chaotic time series, given the previous points $s(n-1),s(n-2),...$. 

After the training, the readout weights $w_\text{out}$ are fixed and the teacher signal $d(n)$ is replaced by the output signal $y(n)$, so that the reservoir becomes autonomous. The evolution of the reservoir computer during the autonomous run is given by Eqs. \ref{eq:rc}, \ref{eq:rcout},  \ref{eq:ucst}  supplemented by
\begin{equation}
  d(n)=y(n) \hspace{1cm} \text{(during autonomous run)}.
  \label{eq:rcAut}
\end{equation}
The reservoir now uses its estimates of the previous points in the time series to estimate the next point.

\section{Training on the Mackey-Glass and Lorenz systems}
\label{sec:carriers}

For illustrative purposes in this work, we use the one-dimensional Mackey-Glass (MG) delay equation and the tri-dimensional Lorenz system. The prediction of the MG and Lorenz systems time series using echo state networks and variants thereof has been investigated previously in a number of works, see \cite{jaeger2004harnessing,shi2007support,li2012chaotic,antonik2017brain,lu2017reservoir,pathak2017using,pathak2018model}.

The Mackey-Glass delay differential equation
\begin{equation}
  \frac{dx}{dt} = \beta \frac{x(t-\tau)}{1+x^n(t-\tau)} - \gamma x,
  \label{eq:mg}
\end{equation}
with $\tau$, $\gamma$, $\beta$, $n > 0$ was introduced to illustrate the appearance of complex dynamics in physiological control systems \cite{mackey1977oscillation}. To obtain chaotic dynamics, we set the parameters as in \cite{jaeger2004harnessing}: $\beta = 0.2$, $\gamma = 0.1$, $\tau = 17$ and $n=10$. With these settings, the highest Lyapunov exponent is $\lambda = 0.006$ \cite{jaeger2004harnessing}. 

Eq. \ref{eq:mg} was integrated using Matlab's \pf{dde23} solver with the initial condition $x(t\leq 0)=0.5$ and integration step of $0.5$ for $7000$ timesteps. The first $1000$ transient values were discarded and the remaining data was split into $3000$ training and $3000$ test inputs.

For the MG task, we used a reservoir with $N=1500$ neurons, the matrix $W$ was rescaled to a spectral radius of $0.79$, while the vectors $w_\text{in}$, $w_\text{back}$ were not rescaled, and 
we set $\Delta=1$, $C=0.44$, $a=0.9$.

At the training stage we obtained an error of $\text{NMSE} = 3 \e{-9}$. During the free run, the error gradually increases, as the reservoir output signal slowly deviates from the target trajectory on the Mackey-Glass attractor. 
Nevertheless, the system manages to generate the desired output for several hundreds of timesteps with reasonable precision.

The Lorenz equations, a system of three ordinary differential equations 
\begin{subequations}
  \begin{align}
    \frac{dx}{dt} & = \sigma \left( y - x \right), \\
    \frac{dy}{dt} & = -xz + rx - y, \\
    \frac{dz}{dt} & = xy - bz,
  \end{align}
  \label{eq:lz}
\end{subequations}
with $\sigma, r, b > 0$, was introduced as a simple model for atmospheric convection \cite{lorenz1963deterministic}. The system exhibits chaotic behaviour for $\sigma=10, b=8/3$ and $r=28$ \cite{hirsch2003differential}, that we used in this study. This yields a chaotic attractor with the highest Lyapunov exponent of $\lambda=0.906$ \cite{jaeger2004harnessing}. 

The Lorenz Eqs. \ref{eq:lz} were integrated using Matlab's \pf{ode45} routine with an integration step of $0.02$ for $10000$ timesteps. We only used the $x$ coordinate of the chaotic system, which was rescaled by the factor of $0.01$, as in \cite{jaeger2004harnessing}.
The first $1000$ transient values were discarded and the remaining data was split into $6000$ training and $3000$ test inputs. 

For the Lorenz task, we used a reservoir of size $N=1500$. We set the spectral radius of the weight matrix $W$ to $0.97$, the input and feedback weights $w_\text{in}$ and $w_\text{back}$ were rescaled to the interval $[-0.5,0.5]$, and we set  $\Delta=1$, $C=0.44$ and $a=0.9$.
To obtain better results we used a form of ridge regression for this task, namely we added noise drawn from the uniform distribution over $[-10^{-6},10^{-6}]$ in the argument of the $\tanh$ in Eq. \ref{eq:rc}.

We obtained a training error of $\text{NMSE} = 3\e{-8}$. The error is one order of magnitude higher here than in the Mackey-Glass case. This may be due to the fact that the Lorenz system has a higher positive Lyapunov exponent and thus exhibits   stronger chaoticity than the Mackey-Glass system, and/or to the fact that the reservoir computer is expected to emulate the dynamics of a 3-dimensional system given only one dimension (the $x$ coordinate), which is more challenging than the reconstruction of the scalar Mackey-Glass system. 

Note: We use here the traditional notation for echo state networks, MG and Lorenz systems, which means that different meanings are given to the same letters. When it is not clear from the context, we use a subscript $x_{RC}, x_{MG}, x_L$ to differentiate them.

\section{Synchronising trained reservoir computers with the Mackey-Glass and Lorenz systems}

Let $s(n)$ be the time series of the chaotic system with which one wishes to synchronise, with $n$ the discretised time used to integrate the Eqs. \ref{eq:mg}, \ref{eq:lz}. We first train the reservoir to predict the next sample in the time series, as described above. Next we start an autonomous run in which the reservoir follows its own dynamics, given by  Eqs. \ref{eq:rc}, \ref{eq:rcout}, \ref{eq:rcAut}. At time $n=n_0$, we start weakly driving the reservoir with the chaotic time series $s(n)$. That is, its dynamics is given by Eqs. \ref{eq:rc}, \ref{eq:rcout}, supplemented by
\begin{equation}
  d(n)=(1-q) y(n) + q s(n)  \hspace{1cm} \text{(when locked)}.
  \label{eq:rcSync}
\end{equation}
with $0 \leq q \leq 1$. 

\begin{figure*}
  \centering
  \subfigure[]{\includegraphics[width=0.32\textwidth]{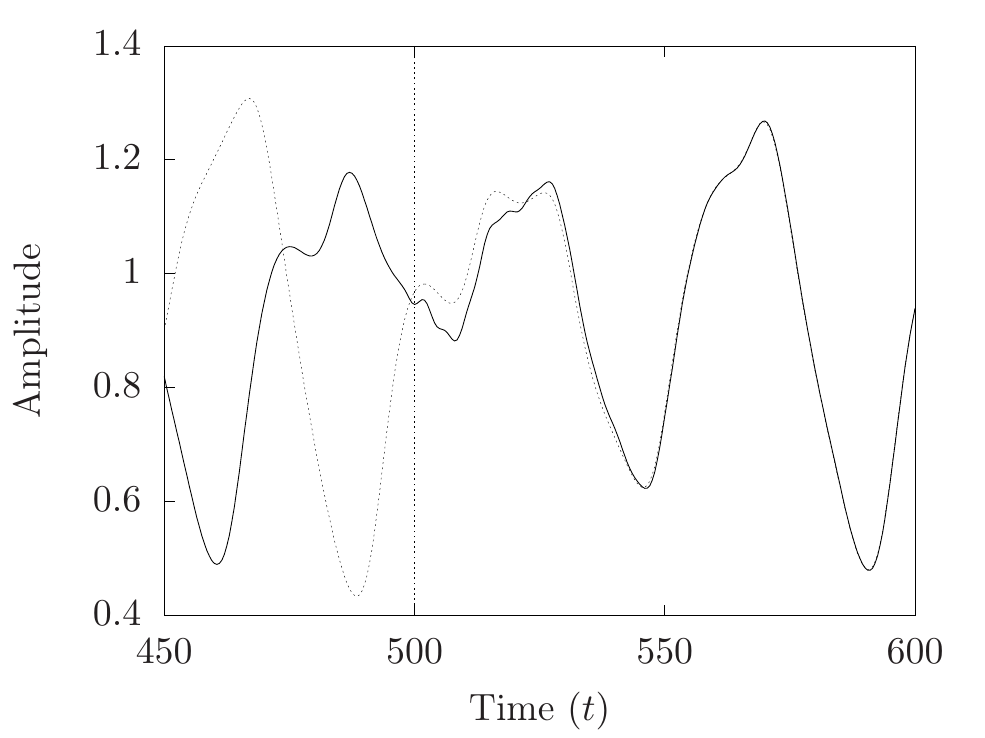}}
  \subfigure[]{\includegraphics[width=0.32\textwidth]{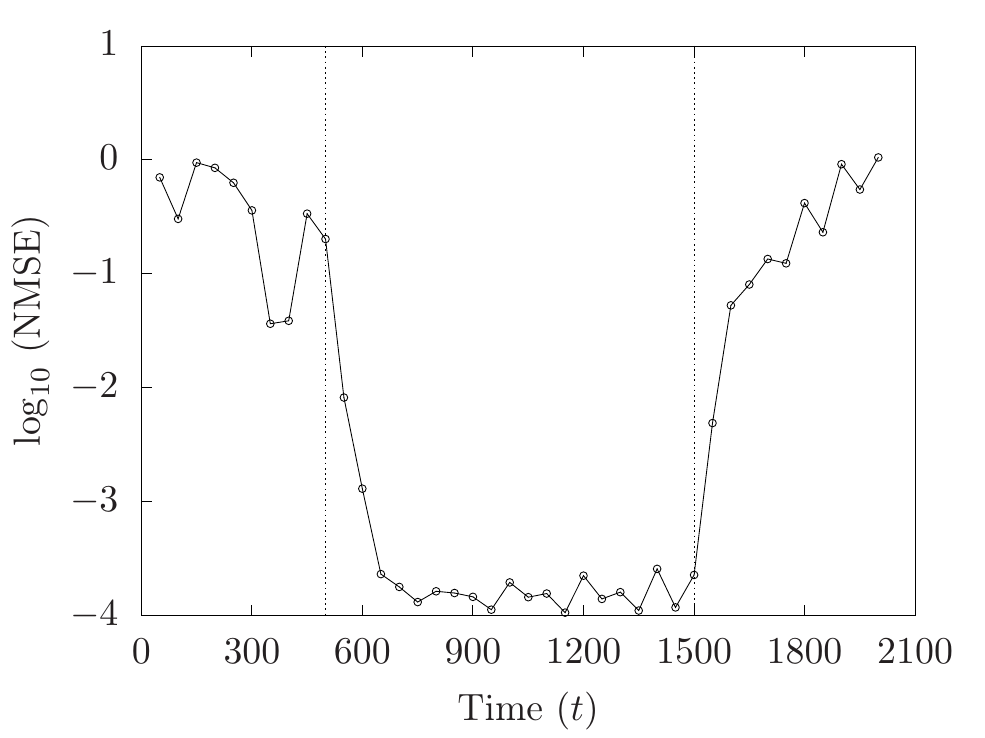}}
 \subfigure[]{\includegraphics[width=0.32\textwidth]{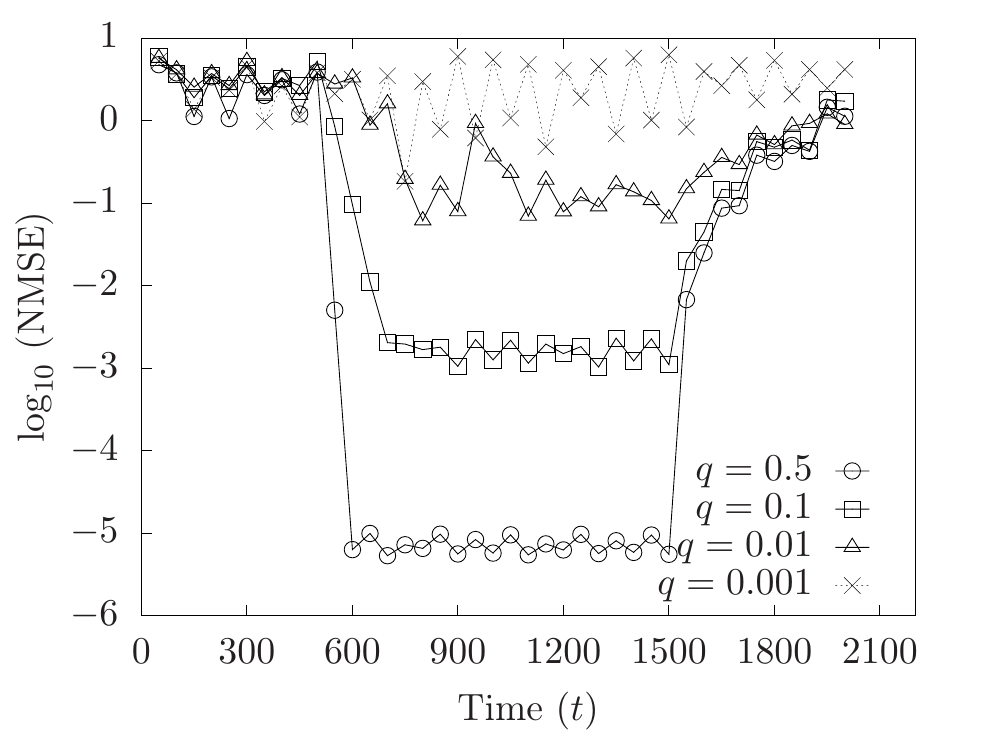}}
  \caption{
    Synchronisation of a trained reservoir computer on the Mackey-Glass system, integrated from the initial condition $x_\text{MG}(t\leq 0)=0.2$. 
    The reservoir computer is trained from $t=-3000$ to $t=0$, whereupon it become autonomous until $t=500$.
    At $t=500$, we set the coupling to $q=0.25$ and remove it ($q=0$) at $t=1500$.
    Plot (a)  depicts only the region of interest around $t=500$, where the reservoir (solid line) synchronises with the chaotic system (dotted line).
    Plot (b) shows the evolution of the NMSE, averaged over 100-timestep intervals, for the entire duration of the simulation, 
    showing the decrease of the NMSE when the synchronisation is turned on, the saturation of the NMSE to a low value ($\text{NMSE}=1.5\e{-4}$) after synchronisation, and the increase of the NMSE when the synchronisation is turned back off.   
    Plot (c) illustrates the same scenario with different coupling strengths. For $q=0.5$, the synchronisation is quicker, as can be seen from the steeper slope, and the resulting NMSE is lower ($\text{NMSE}=7.1\e{-6}$). Lower values of $q$ lead to slower synchronisation, with a higher error. At $q=0.001$, the systems no longer synchronise.
}
  \label{fig:lock1mg}
\end{figure*}

\begin{figure*}
  \centering
  \subfigure[]{\includegraphics[width=0.32\textwidth]{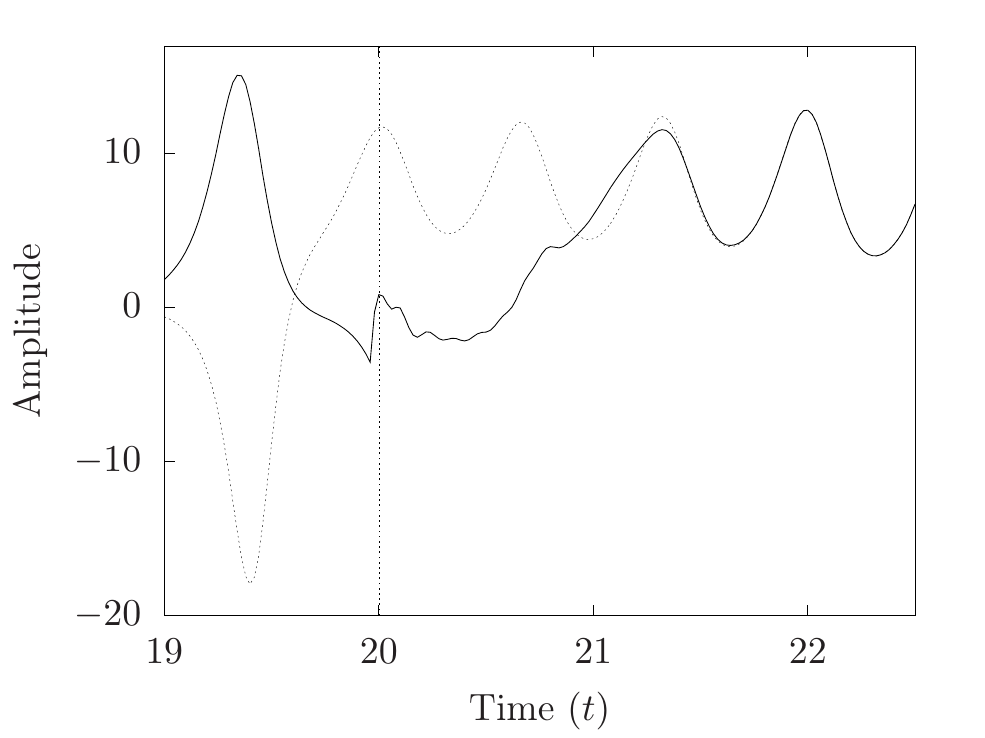}}
  \subfigure[]{\includegraphics[width=0.32\textwidth]{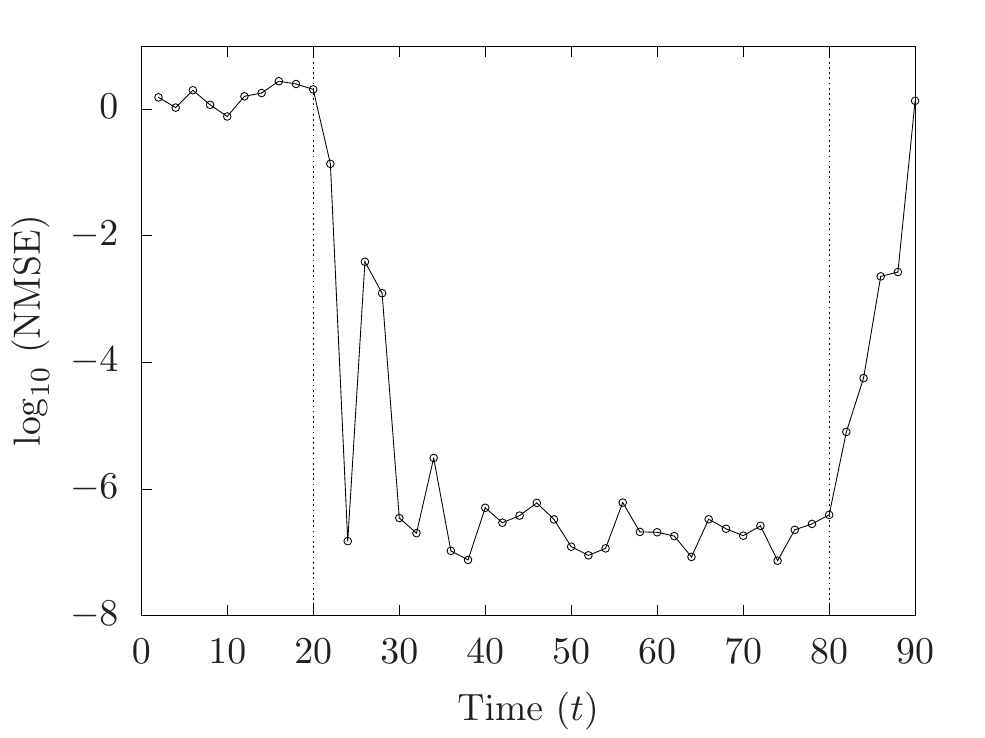}}
  \subfigure[]{\includegraphics[width=0.32\textwidth]{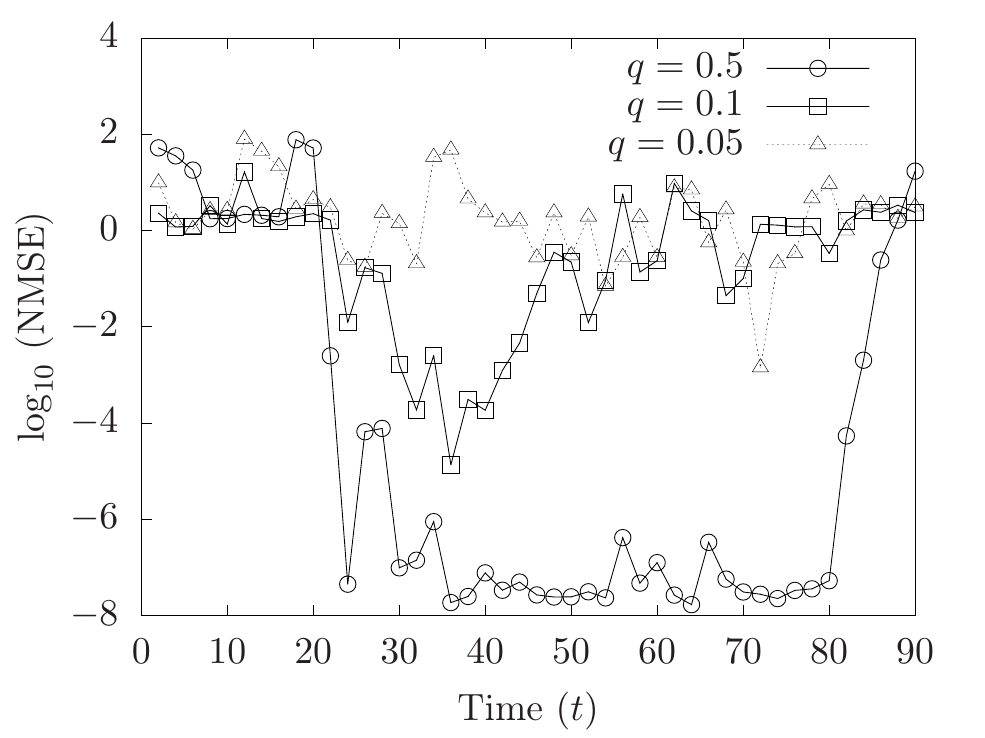}}
  \caption{
    Synchronisation of a trained reservoir computer on the Lorenz system, integrated from $(x_0,y_0,z_0)_\text{LZ}=(10,0,0)$.
    The reservoir computer is trained from $t=-3000$ to $t=0$, whereupon it become autonomous until $t=20$.
    At $t=20$, we set the coupling to $q=0.25$ and remove it ($q=0$) at $t=80$.
    Plot (a)  depicts only the region of interest around $t=20$, where the reservoir (solid line) synchronises with the chaotic system (dotted line).
    Plot (b) shows the evolution of the NMSE, averaged over 100-timestep intervals, for the entire duration of the simulation, showing the decrease of the NMSE when the synchronisation is turned on, the saturation of the NMSE to a low value ($\text{NMSE}=2.3\e{-7}$) after synchronisation, and the increase of the NMSE when the synchronisation is turned back off.   
    Plot (c) illustrates the same scenario with different coupling strengths. Again, higher coupling ($q=0.5$) leads to a lower synchronisation error ($\text{NMSE}=5.1\e{-8}$). Decreasing $q$ leads to a system that tries to synchronise but fails ($q=0.1$). At $q=0.05$, the systems do not synchronise at all.
}
  \label{fig:lock1lz}
\end{figure*}

Figures \ref{fig:lock1mg} and \ref{fig:lock1lz} illustrate how the trained reservoir can lock onto the MG and Lorenz systems. It should be noted that during the synchronisation phase, the NMSE decreases until a minimum value and then stays constant. On the other hand, if we were to synchronise two identical MG or Lorenz systems, the NMSE would decrease until it reached the machine precision. The difference arises because the trained reservoir does not exactly reproduce the dynamics of the MG or Lorenz system.

Taking $q=0.25$, we obtained synchronisation errors of $\text{NMSE}_\text{MG-RC} = 1.5\e{-4}$ and $\text{NMSE}_\text{LZ-RC} = 2.3\e{-7}$. Note that the first subscript corresponds to the primary system, and the seconds indicates the secondary system that is being driven by the primary. Results for other values of $q$ are given in the last panels of Figs. \ref{fig:lock1mg} and \ref{fig:lock1lz}.

\begin{figure*}
  \centering
  \subfigure[]{\includegraphics[width=0.45\textwidth]{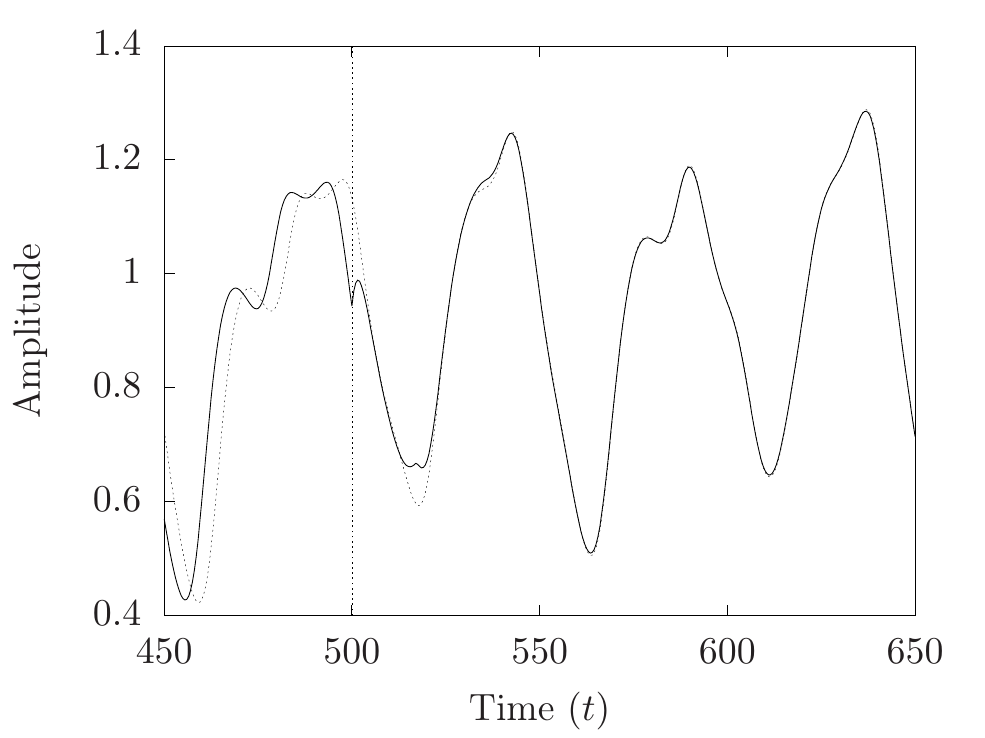}}
  \subfigure[]{\includegraphics[width=0.45\textwidth]{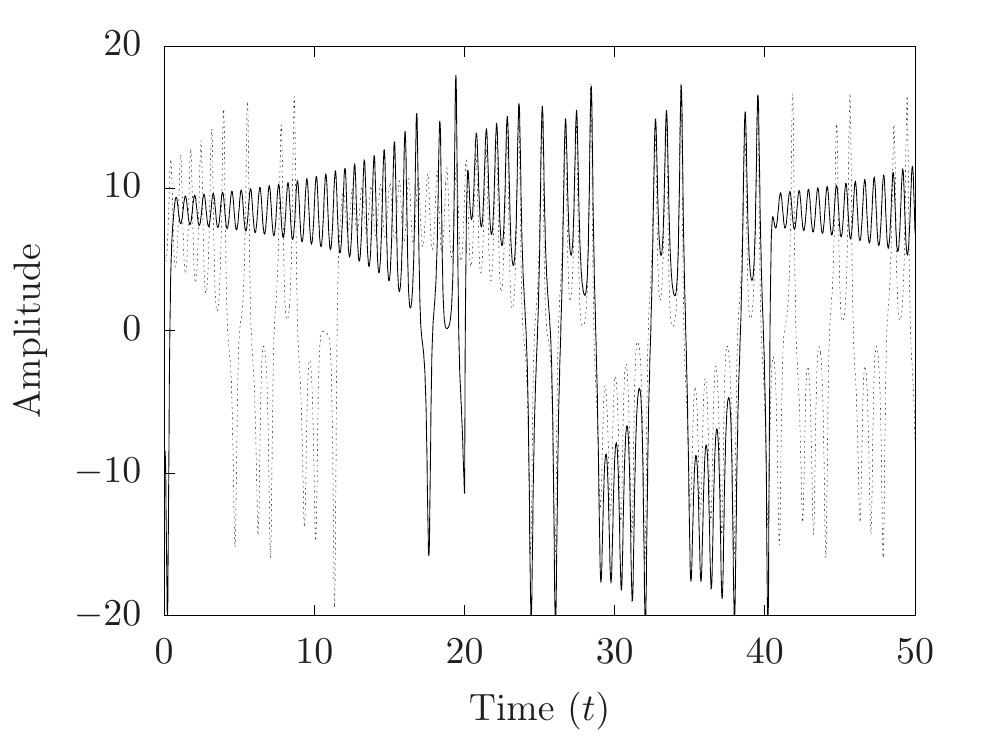}}
  \caption{
    Inverse scenario: synchronisation of (a) Mackey-Glass and (b) Lorenz chaotic systems on a trained reservoir computer.
      The left panel depicts the region of interest around $t=500$, where the Mackey-Glass system (solid lines) synchronises onto the reservoir computer (dotted lines).
      The synchronisation is quite efficient in this case.
      For the Lorenz system, a close view of the region of interest around $t=20$ does not show any interesting dynamics, since the synchronisation error stays quite significant. For this reason, we plot the full time trace instead (right panel) that shows that the synchronised Lorenz system (solid line) accurately follows the switches between the lobes of the Lorenz attractor, emulated by the reservoir computer (dotted lines). The resulting error is much higher in this experiment, but the synchronisation phenomenon can still be observed. At $t=40$ we stopped the synchronisation ($q=0$) and the two systems immediately desynchronised.
}
  \label{fig:lock2}
\end{figure*}

We also tested the inverse scenario, in which a chaotic system (MG or Lorenz) is synchronised on a trained reservoir computer. Let $y_\text{RC}$ be the output of the trained reservoir. In the case of MG, we let Eq. \ref{eq:mg} evolve autonomously until $t=500$. At this time, we change the right hand side of Eq. \ref{eq:mg}, replacing $x(t)$ by
\begin{equation}
  x(t) \rightarrow q y_\text{RC}(t)+ (1-q) x(t).
  \label{eq:synctorc}
\end{equation}
In the case of Lorenz, we let the Eqs. \ref{eq:lz} evolve autonomously until $t=20$, when we change the right hand side of Eqs. \ref{eq:lz}, replacing $x(t)$ by Eq. \ref{eq:synctorc}.
In both cases we took $q=0.25$.
The results are plotted in Fig. \ref{fig:lock2}. We obtained synchronisation errors of $\text{NMSE}_\text{RC-MG} = 1.5\e{-3}$ and $\text{NMSE}_\text{RC-LZ} = 1.5\e{-1}$. Note that the NMSEs are higher than when the RC synchronises onto the MG or LZ systems. This may be due to the fact that the reservoir produces outputs at discrete times separated by the sampling rate $\Delta$, and that this induces a form of noise on the driving signal. We did not investigate in detail the origin of this difference.

\section{Chaos-based cryptography}

Chaos-based cryptography is based on the two ideas that (1) the unpredictable nature of chaotic systems can be used to mask a message and (2) that chaos synchronisation can be used by the receiver to faithfully recover the message. Unfortunately this nice idea has not survived systematic cryptanalysis, basically because the key space (i.e. the parameters describing the chaotic system) is too small, and efficient search methods to recover the key can be developed, see \cite{anstett2006chaotic} and references therein.

The fact that reservoir computers can be trained to emulate chaotic systems, to the extent that the trained reservoir will synchronise with the original chaotic system (as demonstrated in the previous section) suggests that reservoir computing could form the basis for an alternative, conceptually different, approach to cracking chaos based cryptography. 

Here, we demonstrate this on a specific example of how a reservoir computer can be used to crack a chaos based cryptosystem.
We then give some heuristic arguments on why reservoir based approaches could systematically crack chaos based cryptography.
For definiteness, we focus on a scheme previously studied in \cite{prokhorov2008encryption} and referred to as the $\text{III}/1$ scheme. 
This scheme is of interest, because it is similar to many of the systems used in experimental chaos based cryptography that often use delay dynamical systems as chaotic systems, see e.g. \cite{goedgebuer1998optical,argyris2005chaos}.
It has already been cracked in \cite{prokhorov2008encryption}, using time-delay system reconstruction method to recover the unknown parameters of the transmitter. In this work, we use a reservoir computer as an alternative approach. Its advantage over the method in \cite{prokhorov2008encryption} is that the knowledge of the governing equation of the transmitter (see Eq. \ref{eq:mixtransAlice} below) is not required.

We recall the scheme $\text{III}/1$ of \cite{prokhorov2008encryption}, wherein Alice and Bob exchange secret messages, while Eve is eavesdropping.
To encode her message, Alice uses a delay dynamical system in which she injects her message $m(t)$. Her dynamical system obeys the equation
\begin{equation}
  \epsilon \dot{x} (t) = - x(t) + f \left[ x(t-\tau) \right] + m(t), 
  \label{eq:mixtransAlice}
\end{equation}
where $\tau$ is the delay and $\epsilon$ characterises the inertial properties of the system. 
Alice sends $x(t)$ to Bob.

We suppose that $x(t)$ is subject to noise during transmission, so that what is received by Bob is
\begin{equation}
x'(t)=x(t)+\nu(t)
\label{eq:noise}
\end{equation}
where $\nu(t)$ is white noise whose amplitude is given below (i.e. for each successive time point $t$, $\nu(t)$ is independently drawn from the uniform distribution over $[-\nu,\nu]$).

To decrypt the message, Bob uses the same delay system, but in an open loop configuration to obtain the variable $y(t)$ given by
\begin{equation}
  \epsilon \dot{y} (t) = - y(t) + f \left[ x'(t-\tau) \right].
  \label{eq:mixtransBob}
\end{equation}
Then, Bob computes 
\begin{equation}
z'(t) = x'(t) - y(t)
  \label{eq:subxyBob}
\end{equation}
 and obtains an approximate message $m'(t)$ as follows
\begin{equation}
  m'(t) = \epsilon \dot{z'}(t) + z'(t).
  \label{eq:nlm}
\end{equation}
This allows Bob to recover the message $m(t)$, typically corrupted by some high frequency noise. Passing $m'(t)$ through a passband filter centred on the frequency band occupied by $m(t)$ allows Bob to recover a good approximation of $m(t)$.

In order to crack this system, we suppose that Eve has access to a plain text attack, i.e. she has access to both $x'(t)$ and $m(t)$ during some time interval. Thus, she can train her reservoir computer to produce $m(t)$ given $x'(t)$ as input. To this end she uses the scheme described in section \ref{SubSec:TrainCrypto}, in which  the input to the reservoir $u(t)$ is taken to be the signal $x'(t)$ sent by Alice to Bob, and the output of the reservoir ($y(t)$ in Eq. \ref{eq:mse}) is trained to be as close as possible to $m(t)$. 

To illustrate this, we used the MG system Eq. \ref{eq:mg} with the same parameters used elsewhere in this work ($\beta = 0.2$, $\gamma = 0.1$, $\tau = 17$ and $n=10$) (which are identical to the parameters used in  \cite{prokhorov2008encryption} except for $\tau$). 

We first investigate the case  where the  message is a frequency-modulated harmonic signal of the form
\begin{equation}
  m(t) = A \sin \left[ 2 \pi f_c t - B \cos (2\pi f_m t) \right],
  \label{eq:nlmixmsg}
\end{equation}
where $f_c = 5\e{-3}$ is the central frequency of the power spectrum of the signal, $B=3$ is the frequency modulation index, $f_m=5\e{-5}$ is the modulation frequency and $A=0.01$ is the amplitude of the message, chosen to ensure that the information signal comprises $1\%$ of the amplitude of the chaotic carrier. The message and the values of the parameters are  identical to those used in \cite{prokhorov2008encryption}. 
We take the amplitude  of the noise $\nu(t)$ to be $\nu=10^{-1}A$ where $A$ is the maximum amplitude of the message $m(t)$, corresponding to a SNR of $1.5 \times 10^5$.

To crack the system, Eve used a reservoir computer with $N=250$ internal nodes and trained on a plain text message comprising 12000 timesteps.
The spectral radius of the weight matrix $W$ was set to $0.79$, the input weights $w_\text{in}$ were rescaled with a global coefficient of $0.9$, the feedback was switched off $w_\text{back} = 0$, and we set 
$\Delta=0.5$, $C=0.05$ and $a=0.9$.
We obtained a training error of $\text{NMSE} = 3.8\e{-2}$.

\begin{figure*}
  \centering
  \subfigure[]{\includegraphics[width=0.32\textwidth]{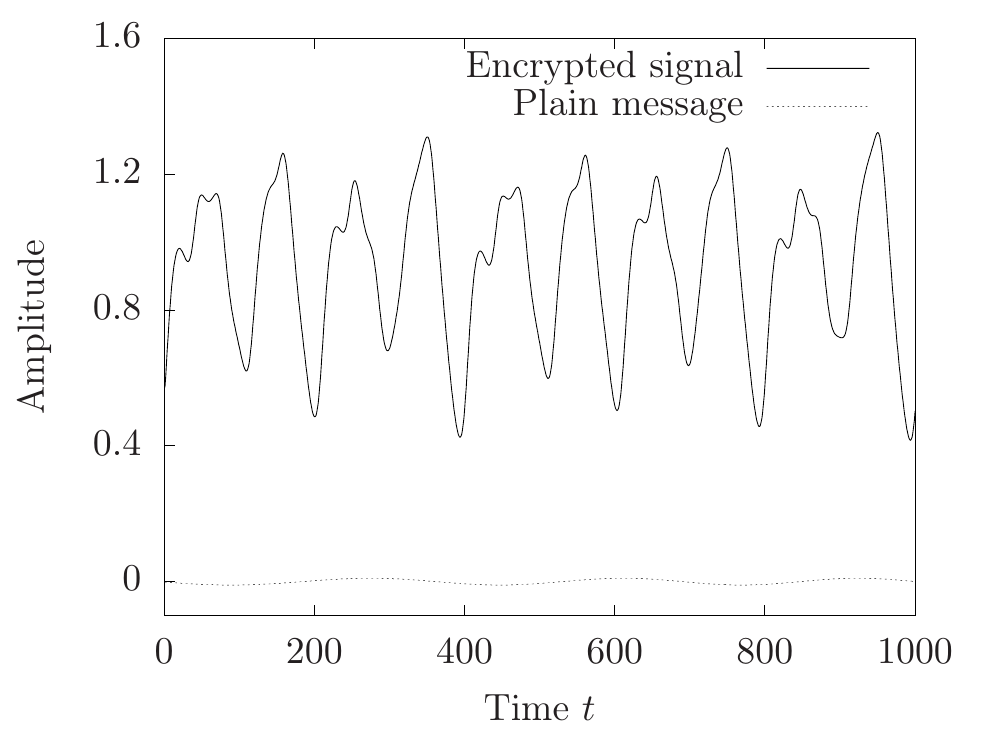}}
  \subfigure[]{\includegraphics[width=0.32\textwidth]{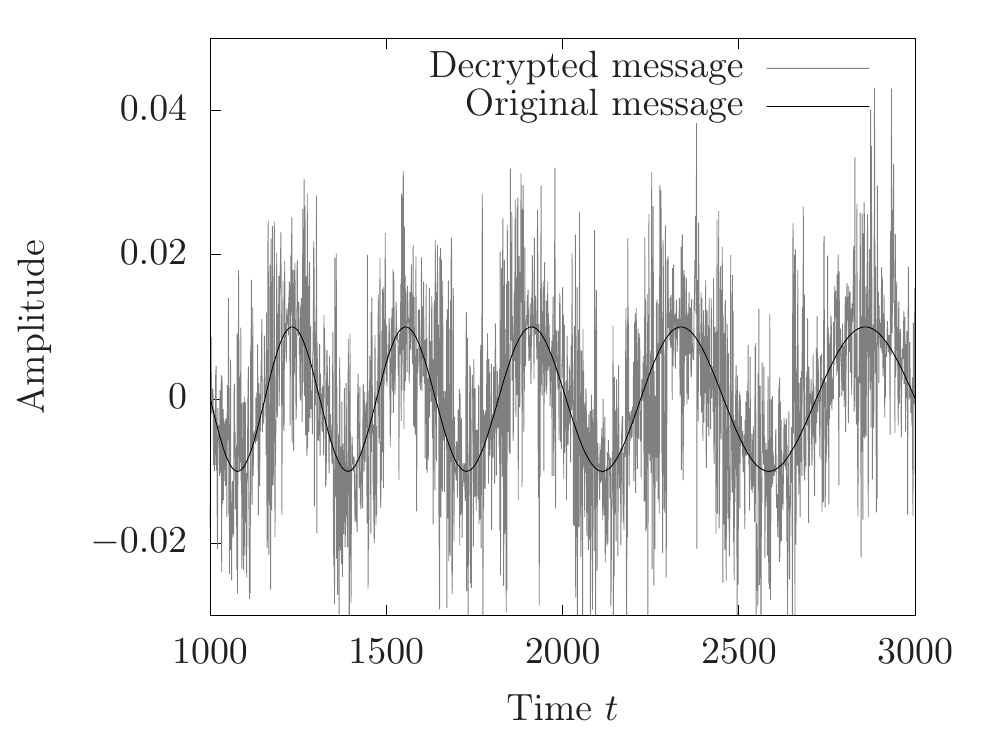}\label{subfig:freqmod_t_bob}}
  \subfigure[]{\includegraphics[width=0.32\textwidth]{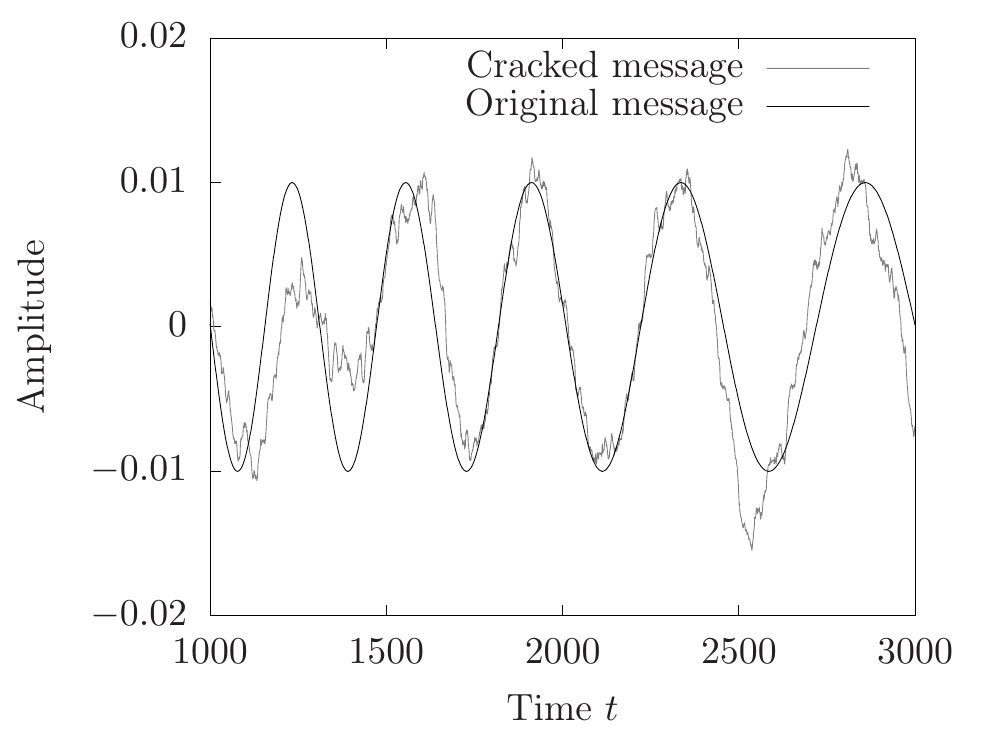}}
  \caption{
    Temporal signals obtained during encryption and decryption of the frequency-modulated sine message Eq. \ref{eq:nlmixmsg}. \textbf{(a)} Encrypted signal $x(t)$ (chaotic carrier + low-amplitude message, solid line) sent by Alice to Bob. The message $m(t)$ (dotted line) is plotted for scale. Note that a small amount of white noise is added to the message during transmission.  \textbf{(b)} Message decrypted by Bob (grey) compared to the original message (black). Despite the high-frequency noise, Bob accurately recovers the frequency modulation of the sine wave. \textbf{(c)} Message obtained by Eve (black) using a trained reservoir computer. Despite slight amplitude variations of the recovered message, Eve accurately recovers the frequency modulation (grey), and with less high-frequency noise than Bob.
}
  \label{fig:nlmix_temp}
\end{figure*}

\begin{figure*}
  \centering
    \subfigure[]{\includegraphics[width=0.45\textwidth]{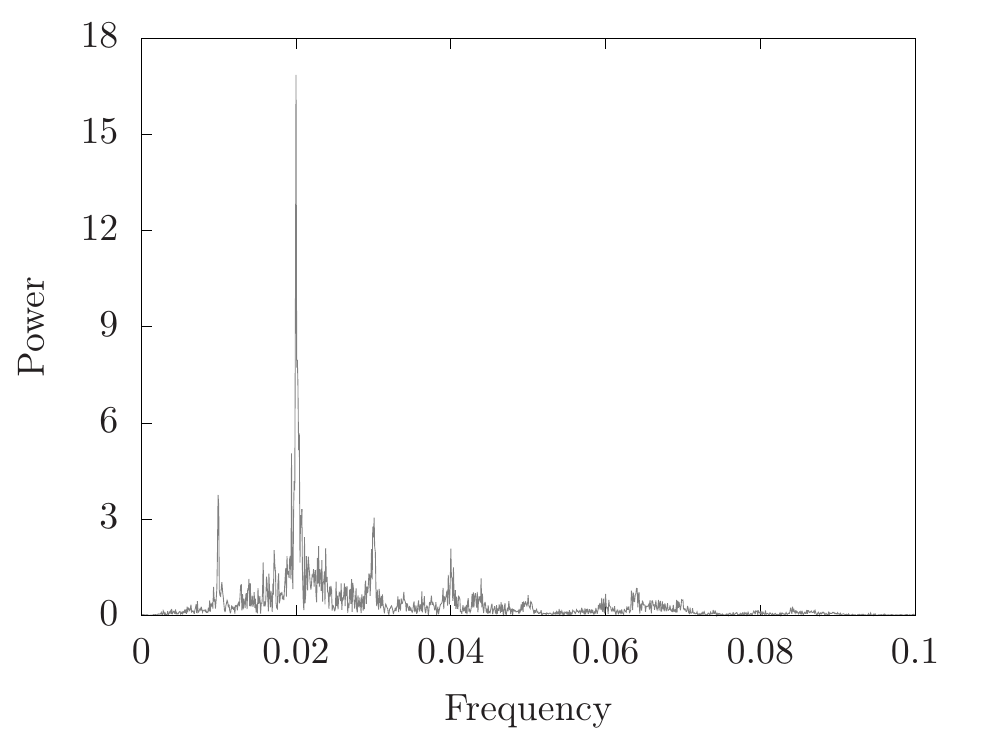}}
    \subfigure[]{\includegraphics[width=0.45\textwidth]{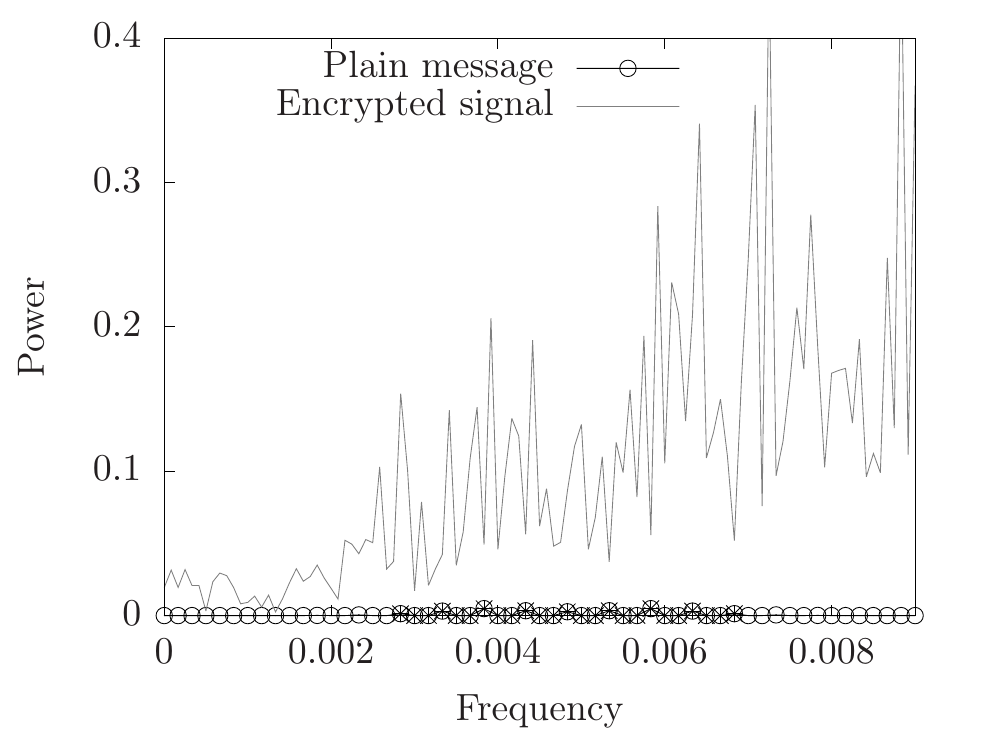}}
  \subfigure[]{\includegraphics[width=0.45\textwidth]{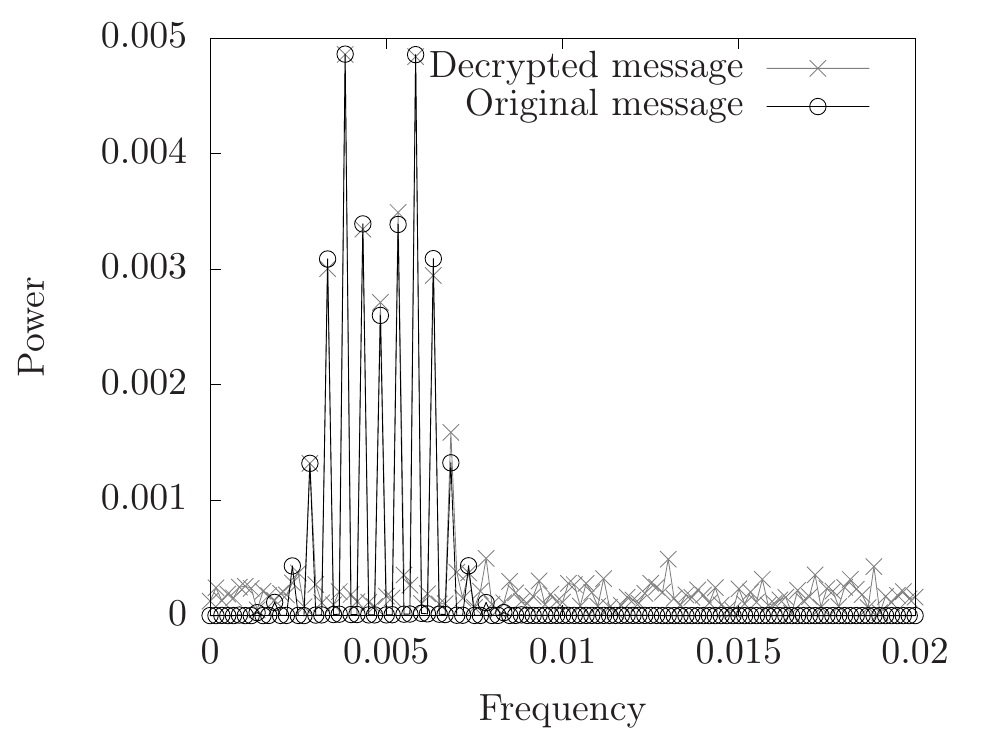}}
  \subfigure[]{\includegraphics[width=0.45\textwidth]{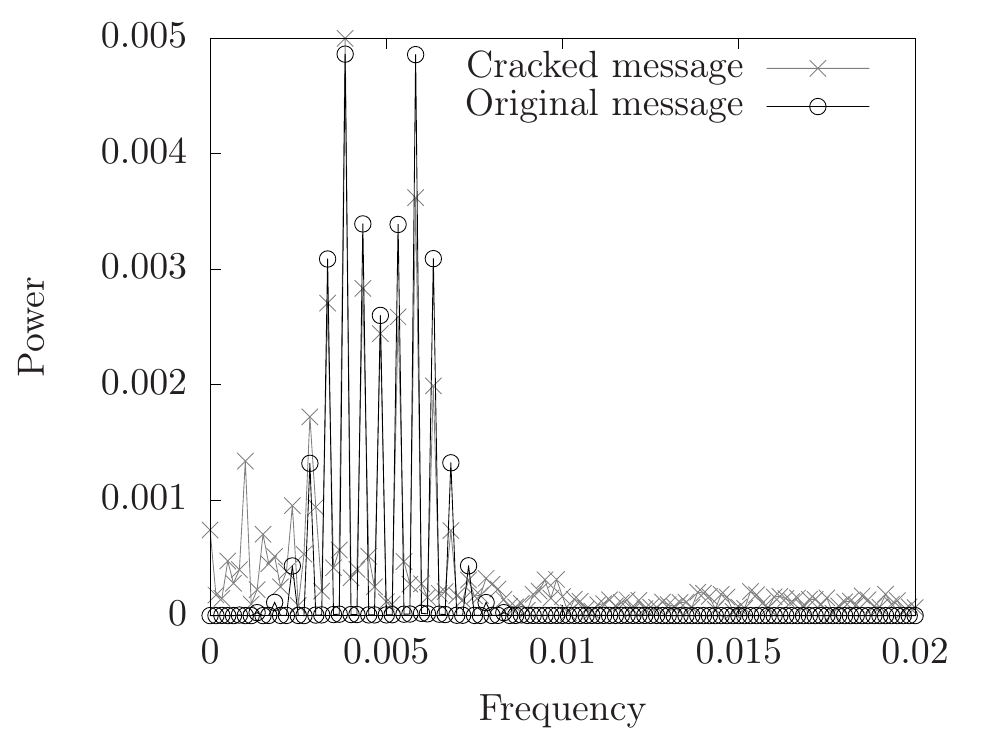}}
  \caption{
    Frequency spectra of signals obtained during encryption and decryption of the frequency-modulated sine message Eq. \ref{eq:nlmixmsg}. 
      \textbf{(a)} Mackey-Glass signal $x(t)$.
      \textbf{(b)} Zoom on the part of the spectrum of the Mackey-Glass signal $x(t)$ which overlaps with the spectrum of the message (the frequency-modulated sine Eq. \ref{eq:nlmixmsg}, highlighted for clarity). Note that the message is well hidden by the chaotic signal $x(t)$.
       \textbf{(c)} Message decrypted by Bob: the frequency modulation is recovered accurately, while the right-hand side of the spectrum (flat, non-zero) corresponds to the high-frequency noise, observed in Fig. \ref{subfig:freqmod_t_bob}.
          \textbf{(d)} Message cracked by Eve: the frequency modulation has been recovered accurately, with a lower level of high-frequency noise.
  }
  \label{fig:nlmix_fft}
\end{figure*}

Using this trained RC, Eve can now try to recover an unknown message sent by Alice. The results are presented in Fig. \ref{fig:nlmix_temp} (temporal signals) and Fig. \ref{fig:nlmix_fft} (frequency spectra), where we compare decryption by Bob and Eve.
Bob manages to accurately retrieve the frequency modulation, but his result is corrupted by high-frequency noise. Therefore, his receiver would benefit from a low-pass filter to get rid of this noise.
Eve's reservoir, on the other hand, incorporates a band-pass filter (the coefficients $a$ and $C$ in Eq. \ref{eq:rc}) that was adjusted to match the frequency band of the message sent by Alice. As a consequence Eve obtains a much cleaner signal. Note that there remain some amplitude variations of Eve's output signal, which may be due to the passband ripples of the low-pass filter of the reservoir computer. However, these ripples do not hinder the retrieval of the frequency modulation. Note that if we set the noise during communication $\nu$ to zero, then Bob's message is of higher quality than Eve's (figures not shown), which is expected since in this case Bob is  carrying out exactly the inverse operation as Alice.

We next investigate a more realistic scenario in which frequency modulation of the sine wave is used to transmit a stream of bits $b(k)$ (with $k\in\mathbb{Z}$) by assigning a higher frequency $\omega_1$ for a ``1'' and a lower $\omega_0$ for a ``0''. In this case, the expression of the encoded message (Eq. \ref{eq:nlmixmsg}) becomes
\begin{equation}
  m(t) = A \sin \left[ \omega_{b(k)} t \right], \quad t\in\left[ kT, k(T+1)\right[,
  \label{eq:nlmix_sinmsg}
\end{equation}
where $T=2\pi/\omega_0$ is the duration of one bit. We take $A=0.02$, $\omega_0 = 0.02 \pi$ and $\omega_1 = 0.04 \pi$. These frequencies are chosen so that the message spectrum is centred on the frequencies where the Mackey-Glass system has largest spectral amplitude, making the system, in principle, harder to crack than the previous example. The amplitude of the noise during transmission is taken to be $\nu=10^{-1} A$, corresponding to a SNR of $3.8 \times 10^4$.

Eve uses a reservoir computer with $N=250$ internal nodes and trained on a plain text message comprising 7000 timesteps.
The spectral radius of the weight matrix $W$ was set to $0.79$, the input weights $w_\text{in}$ were not rescaled, the feedback was switched off $w_\text{back} = 0$, and we set $\Delta=0.1$, $C=0.22$ and $a=0.9$. We obtained a training error of $\text{NMSE} = 2 \times 10^{-1}$.

\begin{figure*}
  \centering
  \subfigure[]{\includegraphics[width=0.45\textwidth]{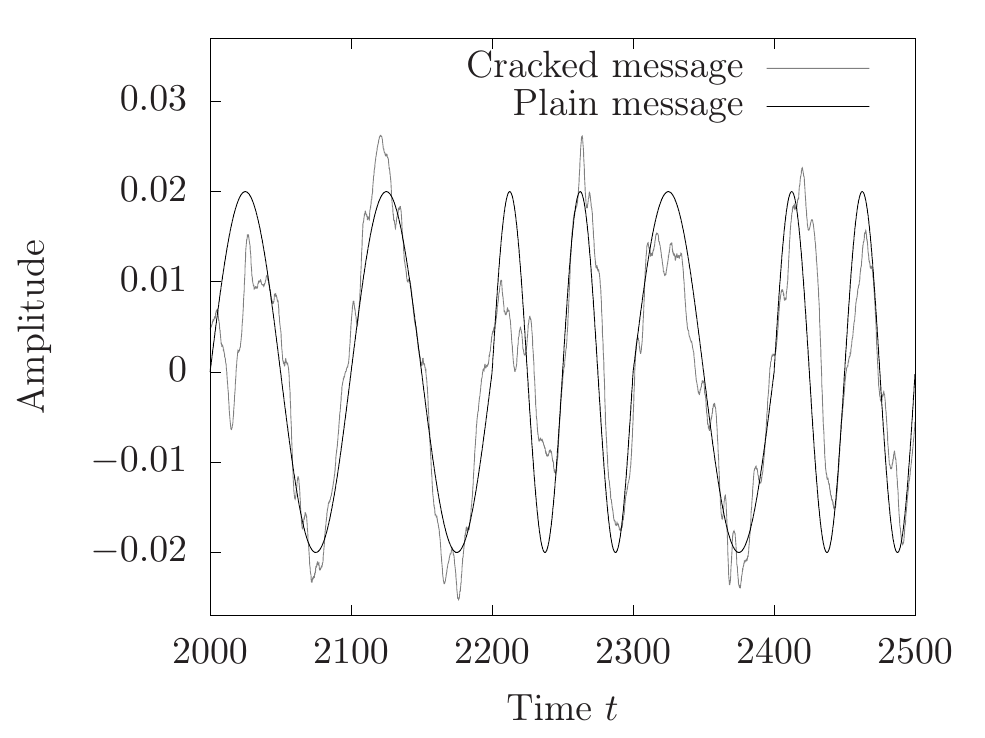}\label{subfig:nlmix_sinseq_temp}}
  \subfigure[]{\includegraphics[width=0.45\textwidth]{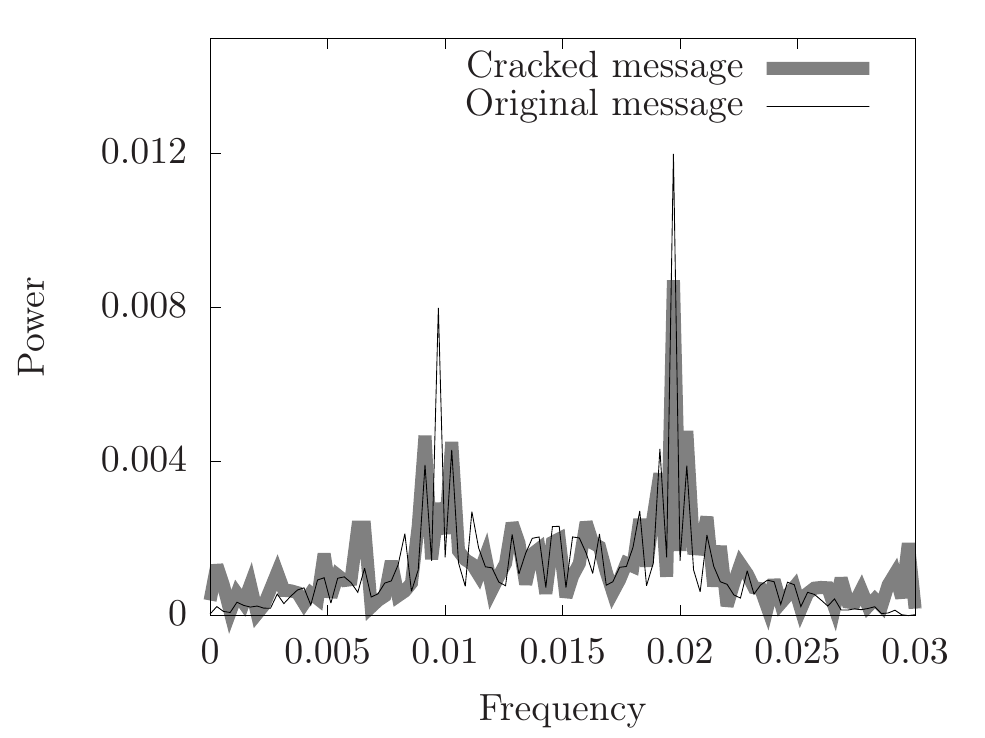}\label{subfig:nlmix_sinmsg_fft}}
  \caption{Decryption of a stream of bits encoded using frequency modulation (Eq. \ref{eq:nlmix_sinmsg}), plotted in \textbf{(a)} time and \textbf{(b)} frequency domains. 
      The sine wave was modulated with frequencies $0.02\pi$ and $0.04\pi$ for bits ``0'' and ``1'', respectively, with a duration of 1 and 2 periods, respectively.
      The cracked signal is not perfectly recovered, but the frequency-coded message can be easily retrieved.
  }
  \label{fig:nlmix_sinmsg}
\end{figure*}

Figure \ref{fig:nlmix_sinmsg} displays the original message (black) and the signal obtained by Eve, using a reservoir computer (grey). Although the recovered signal is not perfect, with significant distortion and noise, one can still accurately recover the  encrypted bit message.

\section{Discussion: why can reservoir computers  emulate chaotic dynamical systems and crack chaos based cryptography ?}

The works \cite{jaeger2004harnessing,antonik2017brain,lu2017reservoir,pathak2017using,pathak2018model}  and the present results show that reservoir computers can be trained to emulate chaotic dynamical systems. Here we try to sketch why this is possible.

The key theoretical concept to understand reservoir computers seems to be the notion of fading memory function \cite{boyd1985fading}, see \cite{dambre2012information,grigoryeva2017universal,marzen2017difference,inubushi2017reservoir}. Consider a time series $u(n)$ with $n$ integer. We denote by $u^{-\infty}(n)=(u(n), u(n-1),u(n-2),...)$ the set of all values up to and including time $n$. Consider a real valued function acting on this left infinite time series $F[u^{-\infty}(n)]$. It has the fading memory property if $F$ depends less and less on $u(n)$ as $n\to-\infty$. 
More precisely, if there is a family of functions $F_m[u(n),u(n-1),...,u(n-m)]$ such that $F_m \to F$ as $m\to \infty$, i.e. we can approximate $F$ by functions of only the last $m$ terms in the time series.

If a reservoir computer obeying the echo state property is driven by a time series $u(n)$, then its internal variables $x_i(n)$ are fading memory functions of the input time series $x_i(n)=X_i[u^{-\infty}(n)]$, and hence the output $y(n)$ of the reservoir computer is also a fading memory function. 

In \cite{grigoryeva2017universal} it was shown that there exists a variant of reservoir computer (based on a polynomial recurrence) that has the following universality property: as the size of the reservoir grows, it can approximate any fading memory function to arbitrary accuracy. This universality property may also hold for reservoirs of the form Eq. \ref{eq:rc}, although this has not been proven. Assuming this hypothesis to be true, given a fading memory function $F[u^{-\infty}(n)]$, a reservoir computer  can approximate $F$ to arbitrary accuracy. That is, the reservoir implements a function $y(n) = F'[u^{-\infty}(n)]$ which is arbitrarily close to $F$, $\vert F'-F\vert\leq \epsilon$ (for an appropriate metric on the space of fading memory functions), where $\epsilon$ can be made arbitrarily small by taking the number of variables of the reservoir $N$ sufficiently large.

Chaotic dynamical systems are also closely linked to the fading memory property. Indeed many chaotic dynamical systems can be expressed as a recurrence of the form
\begin{equation}
d(n)=D[d^{-\infty}(n-1)],
  \label{eq:dD}
\end{equation}
 where $D$ has the fading memory property. This is obviously the case for the logistic map, or for the Mackey-Glass system (upon discretising the time variable), and probably is also true for the Lorenz system (although we have not proven it).

Therefore, if a reservoir computer is driven by such a chaotic time series $d(n)$, it can learn an arbitrarily good approximation $D'$ of the chaotic recurrence. Upon closing the reservoir upon itself, it will obey the recurrence
\begin{equation}
d(n)=D'[d^{-\infty}(n-1)],
  \label{eq:dD'}
\end{equation}
with $\vert D'-D\vert\leq \delta$, where $\delta$ can be made arbitrarily small by taking the number of variables of the reservoir $N$ sufficiently large. One can thus expect that many properties of the original dynamical system will be inherited by the reservoir's emulation. This is confirmed by the numerical studies carried out so far. The above suggests that this could be extended to a formal proof.

We now turn to chaos cryptography. In such a system, Alice sends Bob a times series $s(n)$ in which her message $m(n)$ is hidden. In most such cryptography systems, if not all, Bob's decoding operation will consist of passing the time series $s(n)$ through a fading memory function $M'$ to obtain an approximation $m'(n)$ of the original message $m'(n)=M'[s^{-\infty}(n)]$ where $\vert m' - m\vert <\eta$ for some metric on time series. The fading memory nature of the decoder seems necessary, as it implies that the decoder can act locally on the time series, and does not depend on the values of the time series arbitrarily far in the past. Furthermore, in experimental implementations, the decoding operation must be robust to imperfections. That is, if $M''$ is another fading memory function that is sufficiently close to $M'$ (i.e. $\vert M''-M'\vert< \rho$), then the corresponding decoded message $m''(n)=M''[s^{-\infty}(n)]$ will also be close to the original message if $\rho$ is sufficiently small. 

But this means that given a plain text attack in which the eavesdropper knows the time series $s(n)$ and the corresponding message $m(n)$, she can train a reservoir computer to approximate the fading memory function $M'$. Given the above-mentioned necessary robustness of the cryptographic scheme, if Eve's approximation is good enough, her trained reservoir will now be able to recover the unknown messages.

The above arguments show the plausibility of reservoir computers being able to emulate chaotic systems and to crack chaos-based cryptography. These arguments however say nothing about the efficiency of this approach. Numerical investigations so far suggest that reservoir computers are remarkably good at such emulation tasks. Presumably this is because the reservoir computing approach generates fading memory functions $X_i$ which are in some sense close to the kind of functions produced by natural systems. But of course any other approach that can generate a dense set of fading memory functions (such as, for instance,  Volterra series) will also work, although possibly less efficiently.

\section{Conclusion}

Time series forecasting has been investigated with several different machine learning techniques, in addition to reservoir computing. These include
support vector machines \cite{TAY2001309,KIM2003307}, and auto regressive models and neural networks \cite{ZHANG2003159}.
It would of course be very interesting to compare reservoir computing with other machine learning approaches  for the above tasks of emulating chaotic systems, learning their parameters, chaos synchronisation, cracking chaos cryptography, etc... Such a comparison goes however beyond the present work. 
We expect that reservoir computing will probably report favourably in such a comparison. Indeed as noted above to the best of our knowledge reservoir computers hold the record for predicting the future trajectory of chaotic systems. (Most likely this is because reservoir computers, being recurrent dynamical systems themselves, already encode much of the structure which needed for such a task). An advantage of reservoir computers is that they are particularly easy to train, using only a linear regression.

The present work builds on previous works which showed that reservoir computers with output feedback can emulate chaotic dynamical systems. Previous works focused on forecasting trajectories
\cite{jaeger2004harnessing, grigoryeva2017universal} and predicting spatio-temporal chaos \cite{pathak2018model} , inferring hidden degrees of freedom \cite{lu2017reservoir}, estimating Lyapunov exponents\cite{pathak2017using}. Here we show that trained reservoir computers can synchronise with another chaotic system, thereby demonstrating that the trained reservoir computer has an attractor with similar geometry and stability properties as the original system. We then show how a reservoir computer can be used to crack chaos based cryptography. 

It is interesting to note that cracking chaos based cryptography seems comparatively easy for the reservoir computer. Indeed, while for the time series prediction task (Sec. \ref{sec:carriers}) we used reservoirs with $N=1500$ neurons, for the cryptography application we only used $N=250$ neurons. In addition in the presence of noise in the transmission line the reservoir computer in fact performed better than the system used by Bob. This is in part because the reservoir computer we used comprises a low pass filter.

As noted above, reservoir computers can be implemented in hardware implementations, with good performance and high speed \cite{brunner2012parallel,vandoorne2014experimental,Larger2017high}. The present numerical results suggest that such experimental systems would be good candidates for cracking physically implemented  chaos cryptography.

\section*{Acknowledgements}

We thank Marc Haelterman, Guy Verschaffelt, Guy Van der Sande, and Damien Rontani for helpful discussions. 
This work was supported by the Interuniversity Attraction Poles Program (Belgian Science Policy) Project Photonics@be IAP P7-35, by the Fonds de la Recherche Scientifique (FRS-FNRS), by the Action de Recherche Concert\'ee of the F\'ed\'eration Universitaire Wallonie-Bruxelles through Grant No. AUWB-2012-12/17-ULB9, by the Research Foundation - Flanders (FWO, Ph. D. fellowship), and by the Universit\'e de Namur.


%

\end{document}